\documentclass[letterpaper, 10 pt, conference]{ieeeconf}

\IEEEoverridecommandlockouts
\usepackage{amsmath}
\usepackage{amssymb,bm}
\usepackage{amsfonts}
\usepackage{enumerate}
\usepackage{tabulary}
\usepackage{multirow}
\usepackage{cite}
\usepackage{lipsum} 
\usepackage{verbatim}
\usepackage{hyperref}
\usepackage{url}
\usepackage{balance}
\usepackage{graphicx}
\usepackage{subfigure}
\usepackage{epstopdf}
\usepackage[misc]{ifsym}
\usepackage{color}
\usepackage{xcolor}
\usepackage{xxcolor}
\usepackage{tikz}
\usepackage{pgf}
\usepackage{pgfplots}
\usepackage{float}
\usepackage{ifthen}
\usepackage{forloop}
\usepackage{listings}
\usepackage{booktabs}
\usepackage[lined,boxed,commentsnumbered,linesnumbered]{algorithm2e}

\definecolor{codegreen}{rgb}{0,0.6,0}
\definecolor{codepurple}{rgb}{0.58,0,0.82}
\definecolor{backcolour}{rgb}{0.95,0.95,0.92}
\lstdefinestyle{buzz}{
    backgroundcolor=\color{black!5},   
    commentstyle=\color{codegreen},
    keywordstyle=\color{blue},
    numberstyle=\tiny\color{black!30},
    stringstyle=\color{codepurple},
    basicstyle=\footnotesize\ttfamily,
    breakatwhitespace=false,         
    breaklines=true,                 
    captionpos=b,                    
    keepspaces=true,                 
    numbers=left,                    
    numbersep=5pt,                  
    showspaces=false,                
    showstringspaces=false,
    showtabs=false,                  
    tabsize=2,
}
\SetAlCapSkip{-0.25em}
\title{\LARGE \bf
Learning to Fly---a Gym Environment with PyBullet Physics for Reinforcement Learning of Multi-agent Quadcopter Control
}

\author{Jacopo Panerati,$^{1,2}$ Hehui Zheng,$^{3}$ SiQi Zhou,$^{1,2}$ James Xu,$^{1}$ Amanda Prorok,$^{3}$ and Angela P. Schoellig$^{1,2}$
\thanks{$^{1}$Jacopo Panerati, SiQi Zhou, James Xu, and Angela P. Schoellig are with the \href{http://www.dynsyslab.org}{Dynamic Systems Lab}, Institute for Aerospace Studies, University of Toronto, Canada,
e-mails: {\tt \{name.lastname\}@utoronto.ca};
and the $^{2}$Vector Institute for Artificial Intelligence in Toronto. 
 $^{3}$Hehui Zheng and Amanda Prorok are with the the \href{https://www.proroklab.org}{Prorok Lab} and the Department of Computer Science and Technology, University of Cambridge, Cambridge, United Kingdom, e-mails: {\tt \{hz337, asp45\}@cam.ac.uk}.}
}

\begin{document}
\maketitle
\thispagestyle{empty}
\pagestyle{empty}

\begin{abstract}
Robotic simulators are crucial for academic research and education as well as the development of safety-critical applications.
  Reinforcement learning \emph{environments}---simple simulations
  coupled with a problem specification in the form of a reward 
  function---are also important to standardize the development
  (and benchmarking) of learning algorithms.
  Yet, full-scale simulators typically lack portability and parallelizability.
  Vice versa, many reinforcement learning environments 
  trade-off realism for high sample throughputs
  in toy-like problems. 
While public data sets have greatly benefited deep learning and
  computer vision, we still lack the software tools to
  simultaneously develop---and fairly compare---control theory
  and reinforcement learning approaches.
In this paper, we propose an open-source OpenAI
  Gym-like environment for multiple quadcopters based on the Bullet physics engine.
Its multi-agent and vision-based reinforcement learning interfaces, as well as the support of realistic collisions and aerodynamic effects, make it, to the best of our knowledge, a first of its kind.
We demonstrate its use through several examples, 
  either for control (trajectory tracking with PID control, multi-robot flight with downwash, etc.) or reinforcement 
  learning (single and
  multi-agent stabilization tasks), hoping to inspire future research that combines control theory and machine learning.
\end{abstract}

\section{Introduction}
\label{sec:intro}

Over the last decade, the progress of
machine learning---and deep learning
specifically---has revolutionized
computer science by obtaining (or surpassing) human performance in several tasks, including image recognition and 
game playing~\cite{badia2020}.
New algorithms, coupled with shared benchmarks and data sets have greatly
contributed to the advancement of multiple fields (e.g., as the KITTI suite~\cite{kitti} did for computer vision in robotics).
While reinforcement learning (RL) looks as a very appealing solution to bridge the gap between control theory and deep learning, we are still in the infancy of the creation of tools for the development of realistic continuous control applications through deep RL~\cite{recht2019}.

\begin{figure}[t]
    \centering
    \includegraphics[]{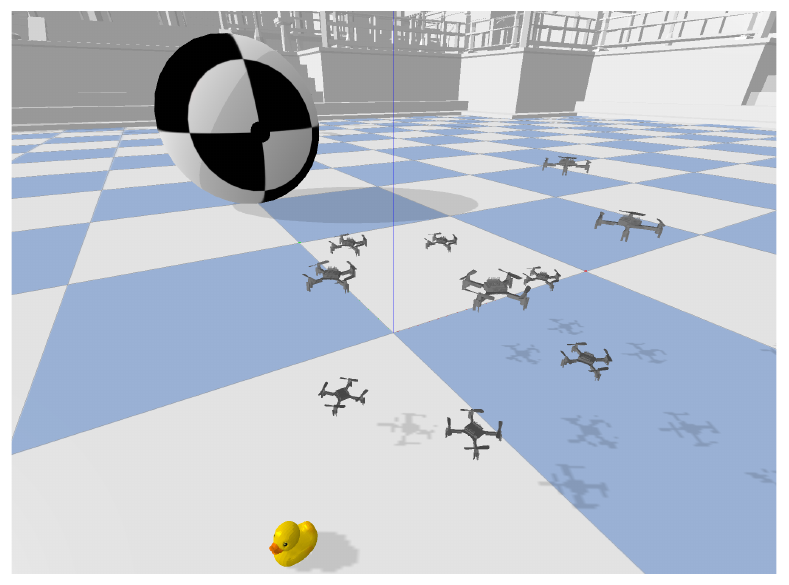}
    \caption{
    Rendering of a \texttt{\scriptsize gym-pybullet-drones} simulation with 10 Crazyflie 2.x on a circular trajectory and a rubber duck for scale.
    }
    \label{fig:page1}
\end{figure}

As automation becomes more pervasive
---from healthcare to aerospace, from package delivery to disaster recovery---better tools to design robotic applications are also required.
Simulations are an indispensable step in
the design of both robots and their
control approaches~\cite{liu2021}, especially so when building 
a prototype is expensive and/or 
safety (of the hardware and its surroundings) is a concern.
Besides platform-specific and proprietary software, many of today's open-source robotic simulators are based on ROS's \emph{plumbing} and engines like Gazebo and Webots.
While these solutions can leverage a host of existing plugins, their limited portability can hinder typical machine learning workflows, based on remote, highly parallel, computing cluster execution.

In an attempt to 
standardize and foster deep RL research, over the last few years, RL \emph{environments} have multiplied.
OpenAI's \emph{Gym}~\cite{brockman2016} emerged as a standard that comprises \emph{(i)}
a suite of benchmark problems as well as \emph{(ii)} an
API for the definition of new environments.
Because of the inherent similarities between the decision-making loops of control theory and RL, many popular \emph{environments} are inspired by control tasks (e.g. the balancing of a pole on a cart).
However, because deep RL algorithms often rely on large amounts of data, some of these environments trade-off realism for high sample throughputs.
A reason for concern is that developing---and 
benchmarking---algorithms on environments that are not necessarily representative of practical scenarios might
curb the progress of RL in robotics~\cite{henderson2018}.

\begin{table*}[ht!]
    \centering
    \caption{
    Feature comparison between this work and recent quadcopter simulators with a focus on RL or the Crazyflie 2.x
    }
    \begin{tabular}{ c  c  c  c  c  c  c  c  c }
    \toprule
& Physics
    & Rendering
    & \multirow{2}{*}{Language}
    & Synchro./Steppable
    & RGB, Depth, and
    & Multiple
    & \multirow{2}{*}{\emph{Gym} API}
    & Multi-agent \\
& Engine
    & Engine
    & 
    & Physics \& Rendering
    & Segmentation Views
    & Vehicles
    & 
    & \emph{Gym}-like API
    \\

    \cmidrule(lr){1-1}
    \cmidrule(lr){2-5}
    \cmidrule(lr){6-7}
    \cmidrule(lr){8-9}

    \textbf{This work} &
    PyBullet & 
    OpenGL3\hyperlink{hyref:01}{$^\dag$} & 
    Python &
    \textbf{Yes} &
    \textbf{Yes} &
    \textbf{Yes} &
    \textbf{Yes} &
    \textbf{Yes}
    \\

    Flightmare~\cite{song2020} &
    \emph{Ad hoc} & 
    Unity & 
    C++ &
    \textbf{Yes} &
    \textbf{Yes} &
    \textbf{Yes} &
    W/o Vision &
    No
    \\

    AirSim~\cite{shah2018} &
    PhysX\hyperlink{hyref:02}{$^\P$} & 
    UE4 &
    C++ &
    No &
    \textbf{Yes} &
    \textbf{Yes} &
    No &
    No
    \\

    CrazyS~\cite{Silano2020} &
    Gazebo\hyperlink{hyref:03}{$^\S$} & 
    OGRE &
    C++ &
    \textbf{Yes} &
    No Segmentation &
    No &
    No &
    No
    \\

    \bottomrule
\end{tabular}
\newline \hfill \newline
{\footnotesize
\hypertarget{hyref:01}{$^\dag$} or TinyRenderer \hspace{1em}
\hypertarget{hyref:02}{$^\P$} or FastPhysicsEngine \hspace{1em}
\hypertarget{hyref:03}{$^\S$} ODE, Bullet, DART, or Simbody
}
\vspace{-2em}     \label{tab:comparison}
\end{table*}

With this work, we want to provide both the robotics
and machine learning (ML) communities with
a compact, open-source \emph{Gym}-style environment\footnote{Video: \color{black} \url{https://youtu.be/VdTsVu1HuYk}}
that supports the definition of multiple learning tasks 
(multi-agent RL, vision-based RL, etc.) on a practical
robotic application: the control of one or more nanoquadcopters.
The software\footnote{\url{https://utiasdsl.github.io/gym-pybullet-drones}}\footnote{\url{https://github.com/utiasDSL/gym-pybullet-drones}}
provided with this paper, \texttt{\small gym-pybullet-drones},
can help both roboticists and ML engineers
to develop end-to-end quadcopter control with model-free or model-based RL.
The main features of \texttt{\small gym-pybullet-drones} are:

\begin{enumerate}
\item \emph{Realism}: support for realistic collisions, aerodynamics effects, and extensible dynamics \emph{via} Bullet Physics~\cite{coumans2019}.
\item \emph{RL Flexibility}: availability of \emph{Gym}-style environments for both vision-based RL and multi-agent RL---simultaneously, if desired.
\item \emph{Parallelizability}: multiple environments can be easily executed, with a GUI or headless, with or without a GPU, with minimal installation requirements.
\item \emph{Ease-of-use}: pre-implemented PID control, as well as Stable Baselines3~\cite{stable-baselines3} and RLlib workflows~\cite{liang2018}.
\end{enumerate}

Section~\ref{sec:related} of this paper reviews similar simulation environments for RL, in general, and quadcopters, specifically.
Section~\ref{sec:methods} details the inner working and programming interfaces of our \emph{Gym} environment.
In Sections~\ref{sec:performance} and~\ref{sec:examples}, we analyze its computing performance and provide control and learning use cases.
Section~\ref{sec:future} suggests the possible extensions of this work. Finally, Section~\ref{sec:conclusions} concludes the paper.

 \section{Related Work}
\label{sec:related}

Several reinforcement learning environments and quadcopter simulators are already available to the community, offering different features and capabilities.
Here, we briefly discuss \emph{(i)} the current landscape of RL environments, \emph{(ii)} the learning interfaces of
existing quadcopter simulators, and \emph{(iii)} how they compare to 
\texttt{\small gym-pybullet-drones}.

\subsection{Reinforcement Learning Environments}

The OpenAI \emph{Gym} toolkit~\cite{brockman2016} 
was created in 2016 to address the lack of standardization 
among the benchmark problems used in reinforcement learning research
and, within five years, it was cited by over 2000 publications.
Besides the standard API adopted in this work, it comprises 
multiple problem sets.
Some of the simplest, ``Classical control'' and ``Box2D'', are two-dimensional, often discrete action problems---e.g., the swing-up of a pendulum.
The more complex problems, ``Robotics'' and ``MuJoCo'', include
continuous control of robotic arms and legged robots in three-dimensions (Swimmer, Hopper, HalfCheetah, etc.) that are based on the proprietary MuJoCo physics engine~\cite{todorov2012}.

MuJoCo's physics engine also powers DeepMind's \texttt{\small dm\_control}~\cite{tassa2020}.
While \texttt{\small dm\_control}'s environments do not expose the same API as \emph{Gym}, they are very similarly structured and DeepMind's suite includes many of the same articulated-body locomotion and manipulation tasks.
However, because of smoothing around the contacts and other simplifications, even locomotion policies trained successfully with these environments do not necessarily exhibit gaits that would easily transfer to physical robots~\cite{recht2019}.

The need for MuJoCo's licensing also led to the development 
and adoption of open-source alternatives such as Georgia Tech/CMU's DART and Google's Bullet Physics~\cite{coumans2019} (with its Python binding, PyBullet).
Open-source Bullet-based re-implementations of the control and locomotion tasks in~\cite{brockman2016} are also provided in \texttt{\small pybullet-gym}.
Community-contributed \emph{Gym} environments like \texttt{\small gym-minigrid}~\cite{gym_minigrid}---a collection of 2D grid environments---were used by over 30 publications
between 2018 and 2021.

Both OpenAI and Google Research have made recent strides
to include safety requirements and real-world uncertainties in their control and legged locomotion RL benchmarks with
\texttt{\small safety-gym}~\cite{ray2019}
and 
\texttt{\small realworldrl\_suite}~\cite{dulacarnold2020}, respectively.

One of the most popular \emph{Gym} environment for quadcopters is \texttt{\small gymfc}~\cite{koch2019}.
While having a strong focus on the transferability to real hardware, the work in~\cite{koch2019} only addresses the learning of an attitude control loop that exceeds the performance of a PID implementation, using Gazebo simulations.
Work similar to~\cite{koch2019}, training a neural network in simulation for the sim2real stabilization of a Crazyflie 2.x (the same quadcopter model used here), is presented in~\cite{molchanov2019}.
To the best of our knowledge, \texttt{\small gym-pybullet-drones} is the first general purpose multi-agent \emph{Gym} environment for quadcopters.

\subsection{Quadcopter Simulators}

RotorS~\cite{Furrer2016} is a popular quadcopter simulator
based on ROS and Gazebo. It includes multiple AscTec multirotor models and simulated sensors (IMU, etc.).
However, it does not come with ready-to-use RL interfaces and its dependency on Gazebo can make it ill-advised for parallel execution or vision-based learning applications. 
CrazyS~\cite{Silano2020}
is an extension of RotorS that is specifically targeted to the Bitcraze Crazyflie 2.x nanoquadcopter.
Due to its accessibility and popularity in research, we also chose the Crazyflie to be the default quadcopter model in \texttt{\small gym-pybullet-drones}.
However, for RL applications, CrazyS suffers from the same limitations as RotorS.

Microsoft's AirSim~\cite{shah2018} is one of the 
best known simulators supporting multiple vehicles---car and quadcopters---and photorealistic rendering through Unreal Engine 4.
While being an excellent choice for the development of
self-driving applications, its elevated computational requirements 
and overly simplified collisions---using FastPhysicsEngine in multirotor mode---make it less than ideal for learning control.
AirSim also lacks a native \emph{Gym} interface, yet wrappers for velocity input control have been proposed~\cite{krishnan2019}.

The most recent and closely related work to ours is ETH's Unity-based Flightmare~\cite{song2020}.
This simulator was created to simultaneously provide photorealistic rendering and very fast, highly parallel dynamics.
Flightmare also implements \emph{Gym}'s API and it includes a single agent RL workflow.
Unlike \texttt{\small gym-pybullet-drones}, however, Flighmare does not include a \emph{Gym} with vision-based observations nor one compatible with multi-agent reinforcement learning (MARL).
Table~\ref{tab:comparison} summarizes the main features of CrazyS, AirSim, Flightmare, and \texttt{\small gym-pybullet-drones}.

 \section{Methods}
\label{sec:methods}

\begin{figure}[t]
    \centering
    \includegraphics[]{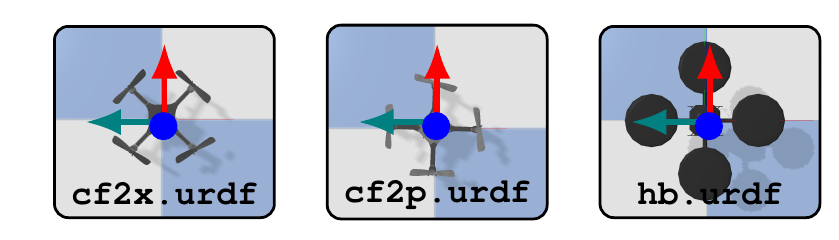}
    \includegraphics[]{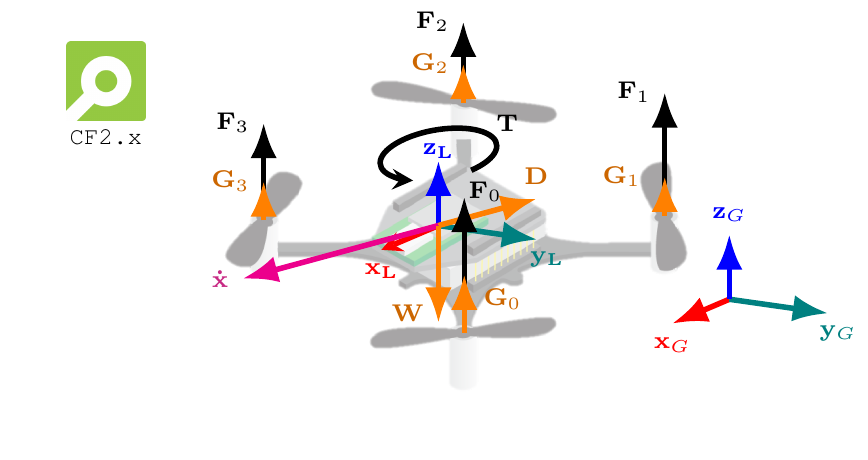}
    \vspace{-3em}
    \caption{
    The three---1 in $\times$-configuration and 2 in $+$-configuration---quadcopter models in \texttt{\scriptsize gym-pybullet-drones} (top) and 
    the forces and torques acting on each vehicle, as modeled in Section~\ref{subsec:dyn} (bottom).
    }
    \label{fig:dynamics}
\end{figure}

To explain how \texttt{\small gym-pybullet-drones} works, one needs to understand 
\emph{(i)} how its dynamics evolve (Subsections~\ref{subsec:gym} to~\ref{subsec:dyn}),
\emph{(ii)} which types of observations, including vision and multi-agent ones, can be extracted from it (Subsection~\ref{subsec:obs}),
\emph{(iii)} what commands one can issue (Subsection~\ref{subsec:act}),
and \emph{(iv)} which learning and control workflows we built on top of it (Subsections~\ref{subsec:workflow} and~\ref{subsec:ros}).

\subsection{Gym Environment Classes}
\label{subsec:gym}

OpenAI's \emph{Gym} toolkit was introduced to 
standardize the development of RL problems and algorithms in Python.
As in a standard Markov decision process (MDP) framework, 
a generic \emph{environment} class receives an
\emph{action}, uses it to update (\emph{step}) its internal state,
and returns a new state (\emph{observation}) paired 
with a corresponding \emph{reward}. 
This, of course, is akin to a simple feedback loop 
in which a controller feeds an input signal to a plant to receive a (possibly noisy) output measurement.
An OpenAI \emph{Gym} environment also exposes additional information about whether its latest state is terminal (\emph{done}) and other standard APIs to \emph{(i)} \emph{reset} 
it between episodes and \emph{(ii)} query it about the domains (\emph{spaces}) of its actions and observations.

\subsection{Bullet Physics}
\label{subsec:bullet}

Physics engines are particularly appealing to researchers
working on both robotics and ML because they \emph{(i)} expedite the development and test of new applications for the former, while \emph{(ii)} yielding large data sets for the latter~\cite{liu2021}. 
The work in this paper is based on the open-source Bullet Physics engine~\cite{coumans2019}.
Our choice was motivated by
its collision management system, the availability of both
CPU- and GPU-based rendering, the compatibility with the Unified Robot Description Format (URDF), and its 
steppable physics---allowing to extract synchronized rendering and kinematics, as well as to control, at arbitrary frequencies.

\subsection{Quadcopter Dynamics}
\label{subsec:dyn}

\begin{figure}[t]
    \centering
    \includegraphics[]{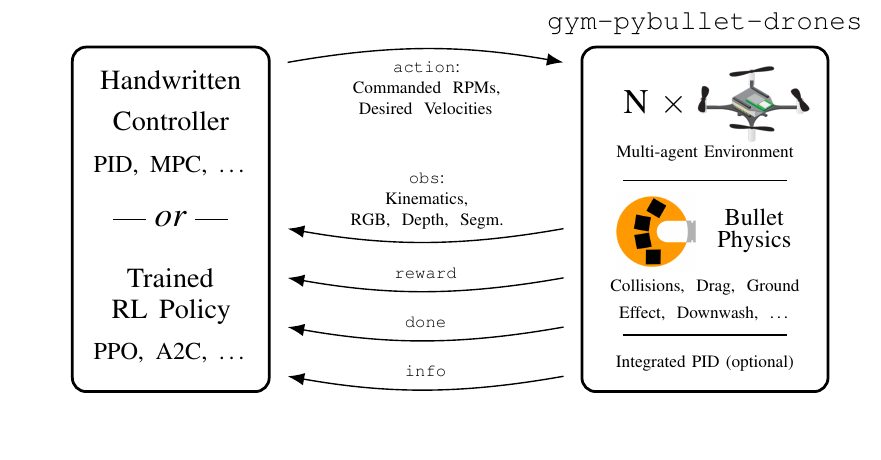}
    \vspace{-1.5em}
    \caption{
    Schematics of the handed over input parameters (\texttt{\scriptsize action}) and yielded return values (\texttt{\scriptsize obs}, \texttt{\scriptsize reward}, \texttt{\scriptsize done}, and \texttt{\scriptsize info}) by every call to the \texttt{\scriptsize step} method of a \texttt{\scriptsize gym-pybullet-drones} environment.
    }
    \label{fig:architecture}
\end{figure}
  
We use PyBullet to model the forces and torques acting 
on each quadcopter in our \emph{Gym} and leverage the physics engine 
to compute and update the kinematics of all vehicles.

\subsubsection{Quadcopter models}

The default quadcopter model in \texttt{\small gym-pybullet-drones}
is the Bitcraze Crazyflie 2.x.
Its popularity and availability meant 
we could leverage both a wealth of system identification
work~\cite{luis2016,forster2015,landry2015} and 
real-world experiments to pick the parameters 
used in this section.
Its arm length $L$, mass $m$, inertial properties $\mathbf{J}$, physical constants, and a convex collision shape are described through separate URDF files for the $\times$ and $+$ configurations (see Figure~\ref{fig:dynamics}).
We also provide the URDF for a generic, larger quadcopter based 
on the Hummingbird described in~\cite{Powers2015}.

\subsubsection{PyBullet-based Physics Update}

First, PyBullet let us set the gravity acceleration $g$
and the physics stepping frequency (which can be much finer
grained than the control frequency at which the \emph{Gym} steps).
Besides the physical properties and constants, PyBullet
uses the URDF information to load a CAD model of the quadcopter.
The forces $F_i$'s applied to each of the 4 motors and the torque $T$
induced around the drone's $z$-axis are proportional to
the squared motor speeds $P_i$'s in RPMs. These are linearly related to the input PWMs and we assume we can control them near-instantaneously~\cite{Powers2015}:
\begin{equation}
    F_{i} = k_F \cdot P_{i} ^ 2 
    , \quad \quad
    T = {\sum}_{i=0}^{3} (-1)^{i+1} k_T \cdot P_{i} ^ 2
    ,
\label{eq:physics}
\end{equation}
where $k_F$ and $k_T$ are constants.

\paragraph{Explicit Python Dynamics Update}

As an alternative implementation, we also provide an explicit Python update that is not based on Bullet. This can be used for comparison, debugging, or the development of \emph{ad hoc} dynamics~\cite{song2020}.
In this case, the linear acceleration in the global frame $\mathbf{\ddot{x}}$ and change in the turn rates in the local frame $\boldsymbol{\dot{\psi}}$ are computed as follows:
\begin{equation}
    \mathbf{\ddot{x}} = \left( \mathbf{R} \cdot [0,0,k_F{\textstyle\sum}_{i=0}^3 P_{i}^2] - [0,0,m g] \right) m^{-1}
    ,
\label{eq:pddot}
\end{equation}
\begin{equation}
    \boldsymbol{\dot{\psi}} = \mathbf{J}^{-1} \left(
    \kappa_{\times}( L, k_F, k_T, [P_{0}^2, P_{1}^2, P_{2}^2, P_{3}^2])
    - \boldsymbol{\psi} \times \left( \mathbf{J} \boldsymbol{\psi} \right) \right)
    ,
\label{eq:psi}
\end{equation}
where $\mathbf{R}$ is the rotation matrix of the quadcopter and $\kappa_{\times}$, $\kappa_{+}$ are functions to compute the torques induced in the local frame by the motor speeds, for the $\times$ and $+$ configuration.

\subsubsection{Aerodynamic Effects}

\begin{figure}[t]
    \centering
    \includegraphics[]{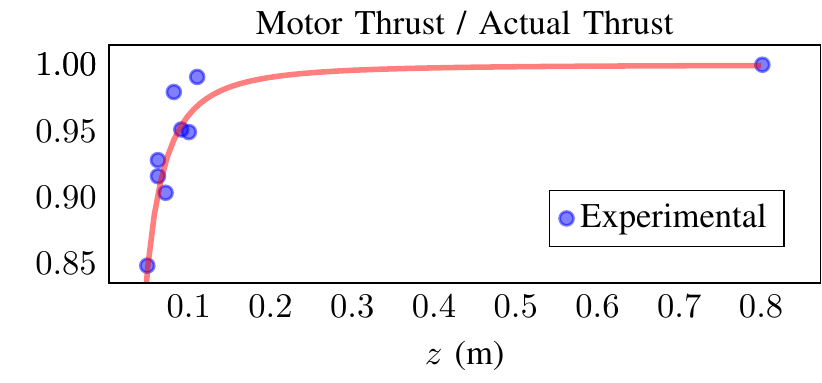}
    \caption{
     The motor thrust-to-actual thrust ratio corresponds to the percentage of thrust actually provided by the motors as a quadcopter hovers at different altitudes in $z$. We used the experimental data in blue to fit the coefficient $k_G$ in \eqref{eq:ground}, yielding the ground effect profile plotted in red. 
    }
    \label{fig:ground}
\end{figure}

\begin{figure*}[t]
    \centering
    \includegraphics[]{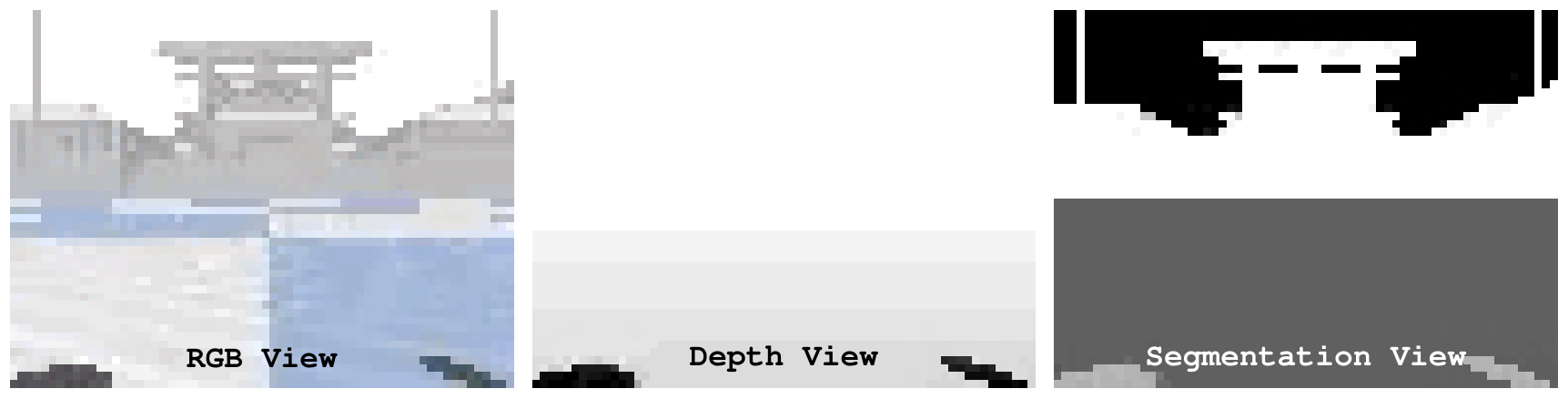} 
    \caption{
    RGB $\mathbf{C}$, depth $\mathbf{U}$, and segmentation $\mathbf{O}$ \texttt{\scriptsize obs} in~\eqref{eq:vision}. A Crazyflie 2.x can be given image processing capabilities by the AI-deck.
    }
    \label{fig:vision}
\end{figure*}

While the model presented in~\eqref{eq:physics}
captures simple quadcopter dynamics, 
flying in a medium, in the proximity of the ground, or near
other vehicles can result in additional aerodynamic effects (Figure~\ref{fig:dynamics}).
PyBullet allows to model these separately and use them jointly.

\paragraph{Drag}

The spinning propellers of a quadcopter produce
drag $\mathbf{D}$, a force acting in the direction
opposite to the one of motion.
Our modeling is based on~\cite{forster2015}
and it states that the air resistance is proportional 
to the quadcopter velocity $\mathbf{\dot{x}}$,
the angular velocities of the rotors, and a matrix of 
coefficients $\mathbf{k_D}$ experimentally derived in~\cite{forster2015}:
\begin{equation}
    \mathbf{D} = - \mathbf{k_D}
    \left(
    {\sum}_{i=0}^{3} \frac{2 \pi P_{i}}{60}
    \right)
    \mathbf{\dot{x}}
    .
\label{eq:drag}
\end{equation}

\paragraph{Ground Effect}

When hovering at a very low altitude,
a quadcopter is subject to an increased thrust 
caused by the interaction of the propellers' airflow with the surface,
i.e., the \emph{ground effect}.
Based on~\cite{shi2019} and real-world experiments with Crazyflie hardware (see Figure~\ref{fig:ground}), we model contributions $G_i$'s for 
each motor that are proportional to the propellers' radius $r_P$,
speeds $P_i$'s, altitudes $h_i$'s, and a constant $k_G$:
\begin{equation}
    G_i =  k_G k_{F} \left( \frac{r_{P}}{4 h_{i} } \right)^2 P_{i}^2
    .
\label{eq:ground}
\end{equation}

\paragraph{Downwash}

When two quadcopters cross paths at different altitudes, the downwash effect causes a reduction in the lift
of the bottom one.
For simplicity, we model it as a single contribution applied to the center of mass of the quadcopter whose module $W$ depends on the distances in $x$, $y$, and $z$ between the two vehicles ($\delta_x$, $\delta_y$, $\delta_z$) and constants $k_{D_1}$, $k_{D_2}$, $k_{D_3}$
that we identified experimentally:
\begin{equation}
    W = k_{D_1} \left( \frac{r_P}{4\delta_z} \right)^2 exp \left( {-\frac{1}{2} \left( \frac{\sqrt{\delta_x^2 + \delta_y^2}}{
    k_{D_2} \delta_z + k_{D_3}}\right)^2} \right)
    .
\label{eq:down}
\end{equation}
Figure~\ref{fig:down} compares a flight simulation using this model with data from a real-world flight experiment.

\subsection{Observation Spaces}
\label{subsec:obs}

Every time we advance \texttt{\small gym-pybullet-drones} by 
one step---which might include multiple steps of the physics engine---we receive an observation vector.
In our code base, we provide several implementations. Yet, they all
include the following kinematic information:
a dictionary whose keys are drone indices $n \in [0 .. N]$ and 
values contain positions $\mathbf{x}_n = [x,y,z]_n$'s,
quaternions $\mathbf{q}_n$'s, rolls $r_n$, pitches $p_n$, and yaws $j_n$'s, linear $\mathbf{\dot{x}}_n$, and angular velocities $\boldsymbol{\omega}_n$'s, as well as the motors' speeds $[P_0,P_1,P_2,P_3]_n$'s
for all vehicles.
\begin{equation}
\{n: [\mathbf{x}_n, \mathbf{q}_n, r_n, p_n, j_n, \mathbf{\dot{x}_n}, \boldsymbol{\omega}_n, [P_{0}, P_1, P_2, P_3]_n] \}
    .
\label{eq:obs}
\end{equation}

\subsubsection{Adjacency Matrix of Multi-Robot Systems}

In networked multi-robot systems, it is convenient to express the notion of \emph{neighboring robots} (those within a certain radius $R$) through adjacency matrix $\mathbf{A} \in \mathbb{R}^{N\times N}$ $A_{ij}$. In this type of observation, the value of each drone's key also includes the drone's corresponding row (of Booleans) in $\mathbf{A}$:
\begin{equation}
    \{ n: \{state: [\dots], neighbors: [A_{n0}, \dots, A_{nN}]\} \}
    .
\label{eq:adj}
\end{equation}

\subsubsection{Vision and Rendering}
  
Furthermore, leveraging PyBullet's bindings to TinyRenderer/OpenGL3 and inspired by Bitcraze's AI-deck, \texttt{\small gym-pybullet-drones} observations can include video frames in each drone's perspective (towards the positive direction of the local $x$-axis) for the RGB ($\mathbf{C}_n \in \mathbb{R}^{64 \times 48 \times 4}$), depth, and segmentation 
($\mathbf{U}_n, \mathbf{O}_n \in \mathbb{R}^{64 \times 48}$)
views (Figure~\ref{fig:vision}).
\begin{equation}
    \{ n: \{\dots, rgb: \mathbf{C}_n, dep: \mathbf{U}_n, seg: \mathbf{O}_n \}  \}
    .
\label{eq:vision}
\end{equation}

\subsection{Action Spaces}
\label{subsec:act}

Advancing a \texttt{\small gym-pybullet-drones} environment by 
one step requires to pass an action (or control input) to it.
For the sake of flexibility, and acknowledging that different
robotic applications (e.g., stabilization vs. path planning) require different levels of abstractions, 
we provide multiple implementations.

\subsubsection{Propellers' RPMs}

The default action space of \texttt{\small gym-pybullet-drones}
is a dictionary whose keys are the drone indices $n \in [0 .. N]$ and 
the values contain the corresponding 4 motor speeds, in RPMs, for each drone:
\begin{equation}
    \{ n: [P_0, P_1, P_2, P_3]_n \}
    .
\label{eq:rpms}
\end{equation}

\subsubsection{Desired Velocity Input}

Alternatively, drones can be controlled
through a dictionary of
desired velocity vectors, in the following format:
\begin{equation}
    \{ n: [v_x, v_y, v_z, v_M]_n \}
    ,
\label{eq:vel}
\end{equation}
where $v_x$, $v_y$, $v_z$ are the components of a unit vector
and $v_M$ is the desired velocity's magnitude.
In this case, the translation of the input into
PWMs and motor speeds is delegated to a PID controller
comprising of position and attitude control subroutines~\cite{Mellinger2011}.

\subsubsection{Other Control Modes}

Developing additional RL and MARL applications
will likely require to tweak and customize observation and 
action spaces.
The modular structure of \texttt{\small gym-pybullet-drones}
is meant to facilitate this.
In Section~\ref{sec:examples}, we provide learning examples
based on one-dimensional action spaces.
The inputs of class \texttt{\small DynAviary} are the desired thrust and 
torques---from which it 
derives feasible RPMs using non-negative least squares.

\subsection{Learning Workflows}
\label{subsec:workflow}
  
Having understood how \texttt{\small gym-pybullet-drones}'s dynamics and observations/actions spaces work,
using it in an RL workflow only requires 
a few more steps.

\subsubsection{Reward Functions and Episode Termination}

Each \emph{step} of an environment should return a 
reward value (or a dictionary of them, for multiple agents).
\emph{Reward} functions are very much task-dependent 
and one must be implemented.
As shown in Section~\ref{sec:examples}, it can be 
as simple as a squared distance.
\emph{Gym}'s \texttt{\small done} and \texttt{\small info} return
values are optional but can be used, e.g., to implement
additional safety requirements~\cite{ray2019}.

\subsubsection{Stable Baselines3 Workflow}

We provide a complete training workflow for single 
agent RL based on Stable Baselines3~\cite{stable-baselines3}.
This is a collection of RL algorithms---including A2C, DDPG, PPO, SAC, and TD3---implemented in PyTorch.
As it supports both MLP and CNN policies, Stable Baselines3 can be used with either kinematics or vision-based observations.
In Section~\ref{sec:examples}, we show how to run a training example and replay its best performing policy.

\subsubsection{RLlib Workflow}

We also provide an example training workflow for multi-agent RL based on RLlib~\cite{liang2018}.
RLlib is a library built on top of Ray's API for distributed applications, which includes TensorFlow and PyTorch 
implementations of many popular RL (e.g., PPO, DDPG, DQN) and MARL (e.g., MADDPG, QMIX) algorithms.
In Section~\ref{sec:examples}, we show how to run a 2-agent centralized critic training example and replay its best performing policies.

\subsection{ROS2 Wrapper Node}
\label{subsec:ros}
  
Finally, because of the significance of ROS for the
robotics community, we also implemented a minimalist 
wrapper for \texttt{\small gym-pybullet-drones}'s 
environments using a ROS2 Python node.
This node continuously steps an environment while \emph{(i)} publishing
its observations on a topic and \emph{(ii)} reading actions from a separate topic it subscribed to.
 \section{Computational Performance}
\label{sec:performance}

\begin{table}[]
    \centering
    \caption{
    CPU and GPU speed-ups as a function of the number vehicles, environments, and the use of vision-based observations
    }
    \begin{tabular}{ c  c  c  c  c }
    \toprule
\# of Drones & \# of Env's & Vision 
    & TinyRenderer\hyperlink{hyref:11}{$^\ddag$}
     & OpenGL3\hyperlink{hyref:12}{$^\parallel$}  \\
    \cmidrule(lr){1-3} \cmidrule(lr){4-4} \cmidrule(lr){5-5}
    $1.0$ &
    $1.0$ &
    No &
    $\mathbf{16.8\times}$ &
    $15.5\times$\\
$1.0$ &
    $1.0$ &
    Yes &
    $1.3\times$ &
    $\mathbf{10.8\times}$\\
$5.0$ &
    $1.0$ &
    Yes &
    $0.2\times$ &
    $\mathbf{2.5\times}$\\
$10.0$ &
    $1.0$ &
    No &
    $2.3\times$ &
    $2.1\times$\\
$80.0$ &
    $4.0$ &
    No &
    $\mathbf{0.95\times}$ &
    $0.8\times$\\
\bottomrule
\end{tabular}
\newline \hfill \newline
{\footnotesize \hypertarget{hyref:11}{$^\ddag$} 2020 MacBook Pro (CPU: i7-1068NG7)} \\
{\footnotesize \hypertarget{hyref:12}{$^\parallel$} Lenovo P52 (CPU: i7-8850H; GPU: Quadro P2000)}

     \label{tab:performance}
\end{table}

To demonstrate our proposal, we first analyze its computational efficiency.
Being able to collect large data sets---through parallel execution and running faster than the wall-clock---is of particular importance for the development of reinforcement learning applications.
We chose to adopt \emph{Gym}'s Python API~\cite{brockman2016}, while leveraging Bullet's C++ back end~\cite{coumans2019}, to strike a balance between readability and portability, on one side, and computational performance, on the other.

\begin{figure}[t]
    \centering
    \includegraphics[]{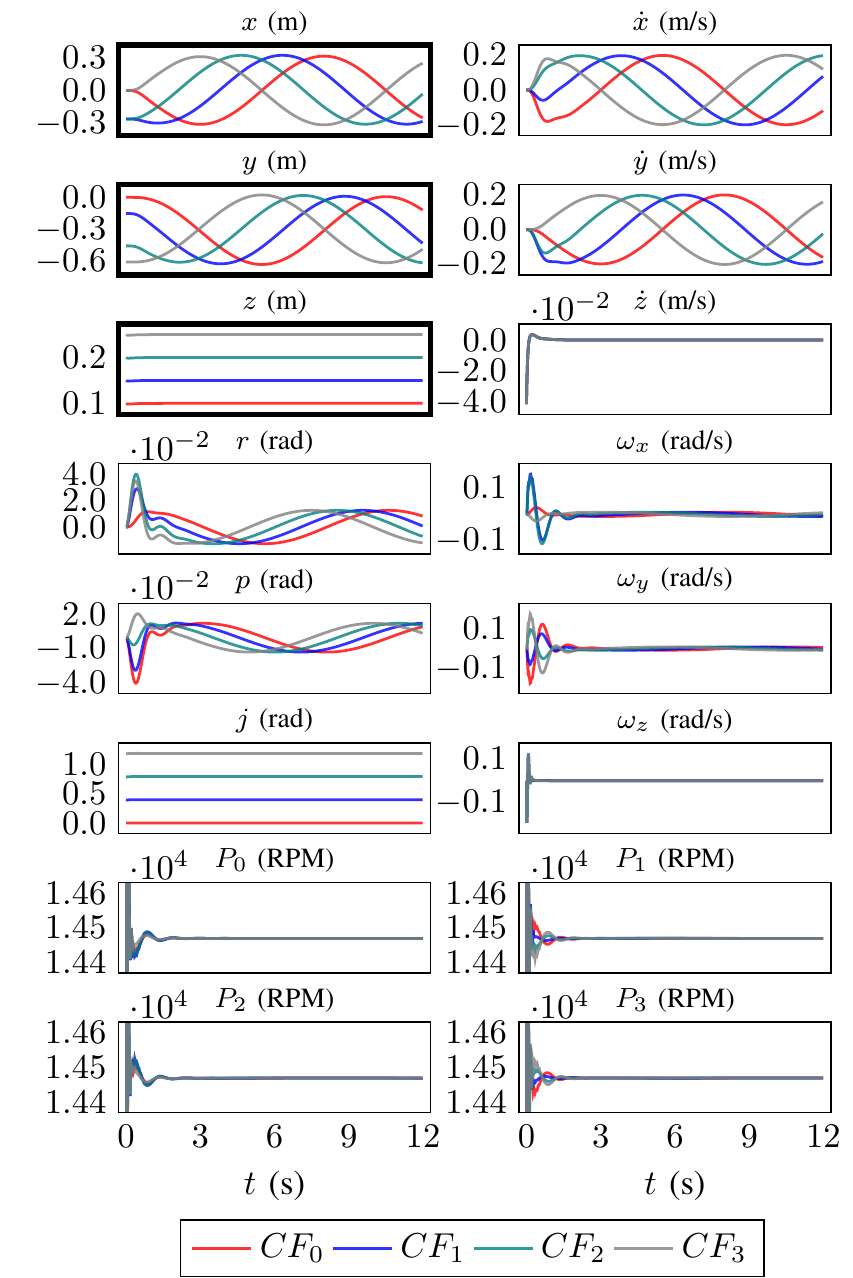}
    \caption{
    Positions in $x$, $y$, $z$, linear velocities $\dot{x}$, $\dot{y}$, $\dot{z}$, roll $r$, pitch $p$, yaw $j$, angular velocities $\boldsymbol{\omega}$, and motors' speeds $P_0$, $P_1$, $P_2$, $P_3$ of four Crazyflies $CF_0$, $CF_1$, $CF_2$, $CF_3$ tracking a circular trajectory, at different altitudes $z$'s, with different yaws $j$'s, \emph{via} external PID control.
    }
    \label{fig:pid}
\end{figure}

As we believe that closed-loop performance is the better gauge of a simulation at work, we used a stripped-down version of the PID control~\cite{Mellinger2011} example in Figure~\ref{fig:pid} to generate the data presented in Table~\ref{tab:performance}.
The script---with no GUI, no debug information, fewer obstacles, and less front end reporting between physics steps---is available in folder:
\texttt{\small gym-pybullet-drones/experiments/performance/}.

Unlike simulations that rely on game engines like Unreal Engine 4 and Unity~\cite{shah2018, song2020}, PyBullet has less demanding rendering requirements and can run with either the CPU-based TinyRenderer or OpenGL3 when GPU acceleration is available.
We collected the results in Table~\ref{tab:performance} using two separate laptop workstations: \emph{(i)} a 2020 MacBook Pro with an Intel i7-1068NG7 CPU and \emph{(ii)} a Lenovo P52 with an Intel i7-8850H CPU and an Nvidia Quadro P2000 GPU.

For a single drone with physics updates at 240Hz, we achieved speed-ups of over 15$\times$ the wall-clock.
Exploiting parallelism (i.e., multiple vehicles in multiple environments), we could generate 80$\times$  the data of the elapsed time.
This is slightly slower, but comparable, to Flightmare's dynamics open-loop performance.
Although on simpler scenes, a visual observations throughput of $\sim$750kB/s with TinyRenderer is also comparable to Flightmare, and 10$\times$ faster when OpenGL3 acceleration is available.

\begin{figure}[t]
    \centering
    \includegraphics[]{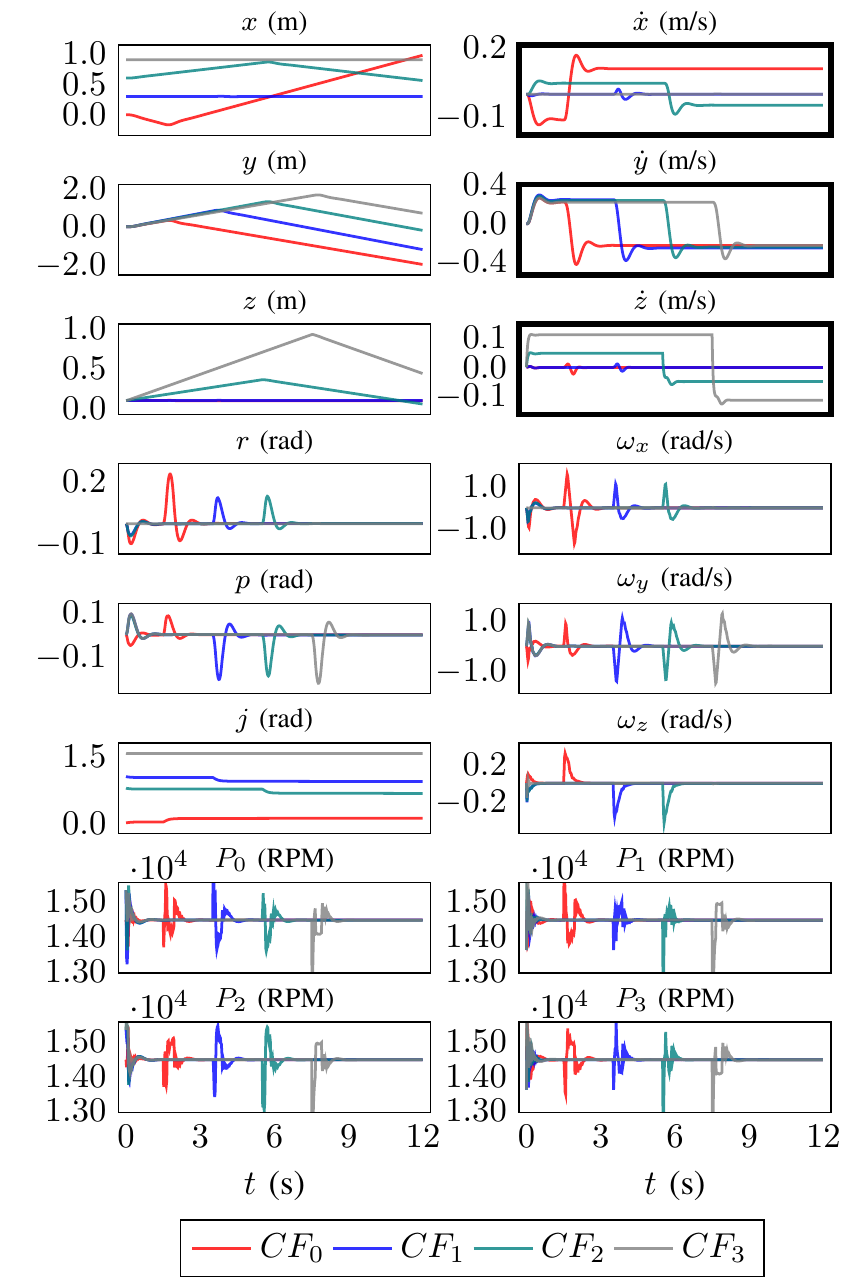}
    \caption{
    Positions in $x$, $y$, $z$, linear velocities $\dot{x}$, $\dot{y}$, $\dot{z}$, roll $r$, pitch $p$, yaw $j$, angular velocities $\boldsymbol{\omega}$, and motors' speeds $P_0$, $P_1$, $P_2$, $P_3$ of four Crazyflies $CF_0$, $CF_1$, $CF_2$, $CF_3$ tracking step-wise
    desired velocity inputs \emph{via} PID control embedded within the \emph{Gym} environment.
    }
    \label{fig:vel}
\end{figure}

\section{Examples}
\label{sec:examples}

In practice, we show how one can jointly use \texttt{\small gym-pybullet-drones} with both control approaches and reinforcement learning algorithms.
We do so through a set of six examples.
All of our source code is available online and it can be installed using the following steps (please refer to the repository for a full list of requirements):
\begin{lstlisting}[firstnumber=1,language=Python,
    numbers=none,
label = {alg:install},]
$ git clone -b paper \
    git@github.com:utiasDSL/gym-pybullet-drones.git
$ cd gym-pybullet-drones/
$ pip3 install -e .
\end{lstlisting}

\subsection{Control}

The first four examples demonstrate how to command multiple quadcopters using motor speeds or desired velocity control inputs as well as two of the aerodynamic effects discussed in Section~\ref{sec:methods}: ground effect and downwash.
\begin{lstlisting}[firstnumber=1,language=Python, 
    numbers=none,
label = {alg:ctrl},]
$ cd gym-pybullet-drones/examples/
\end{lstlisting}

\subsubsection{Trajectory tracking with PID Control}

The first example includes 4 Crazyflies
in the $\times$ configuration, using PyBullet's physics update~\eqref{eq:physics}, and \emph{external} PID control. The controllers receive the
kinematics observations~\eqref{eq:obs} and return
commanded motors' speeds~\eqref{eq:rpms}.
\begin{lstlisting}[firstnumber=1,language=Python, 
    numbers=none,
label = {alg:pid},]
$ python3 fly.py --num_drones 4
\end{lstlisting}
Figure~\ref{fig:pid} plots position $\mathbf{x}$, velocity $\mathbf{\dot{x}} = [\dot{x},\dot{y},\dot{z}] $, roll $r$, pitch $p$, yaw $j$, angular velocity $\boldsymbol{\omega}$, and motors' speeds $[P_0, P_1, P_2, P_3]$ for all vehicles, during a 12 seconds flight along a circular trajectory (the three top-left subplots).
    
\subsubsection{Desired Velocity Input}

The second example also uses kinematics observations~\eqref{eq:obs} but it is controlled by
desired velocity inputs~\eqref{eq:vel}.
These are targeted by PID controllers \emph{embedded} within the environment.
\begin{lstlisting}[firstnumber=1,language=Python, 
    numbers=none,
label = {alg:vel},]
$ python3 velocity.py --duration_sec 12
\end{lstlisting}
Figure~\ref{fig:vel} shows the linear velocity response to
step-wise changes in the velocity inputs (the three top-right subplots).

\subsubsection{Ground Effect}

\begin{figure}[t]
    \centering
    \includegraphics[]{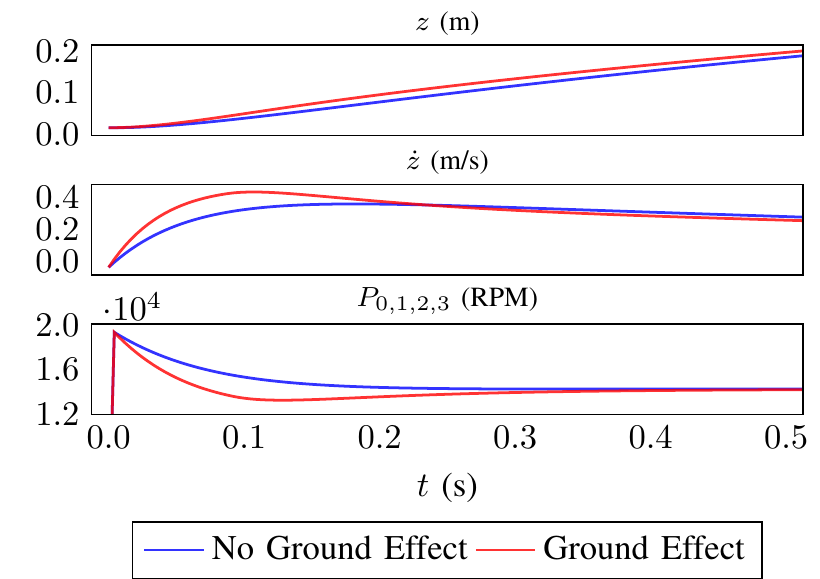}
    \caption{
    Position in $z$, linear velocity $\dot{z}$, and motors' speeds $P_0 = P_1 = P_2 = P_3$ of a simulated Crazyflie taking-off with (red) and without (blue) the modeled ground effect contribution in~\eqref{eq:ground}.
    }
    \label{fig:ground2}
\end{figure}

The third example compares the take-off of a Crazyflie with and without the ground effect contribution~\eqref{eq:ground}, using the coefficient identified in Figure~\ref{fig:ground}.
\begin{lstlisting}[firstnumber=1,language=Python, 
    numbers=none,
label = {alg:ground},]
$ python3 groundeffect.py
\end{lstlisting}
Figure~\ref{fig:ground2} compares positions and velocities along the global $z$-axis, during the first half-second of simulation time,
showing a small but noticeable overshooting and the larger maximum velocity of the drone experiencing the ground effect.

\subsubsection{Downwash}

\begin{figure}[t]
    \centering
    \includegraphics[]{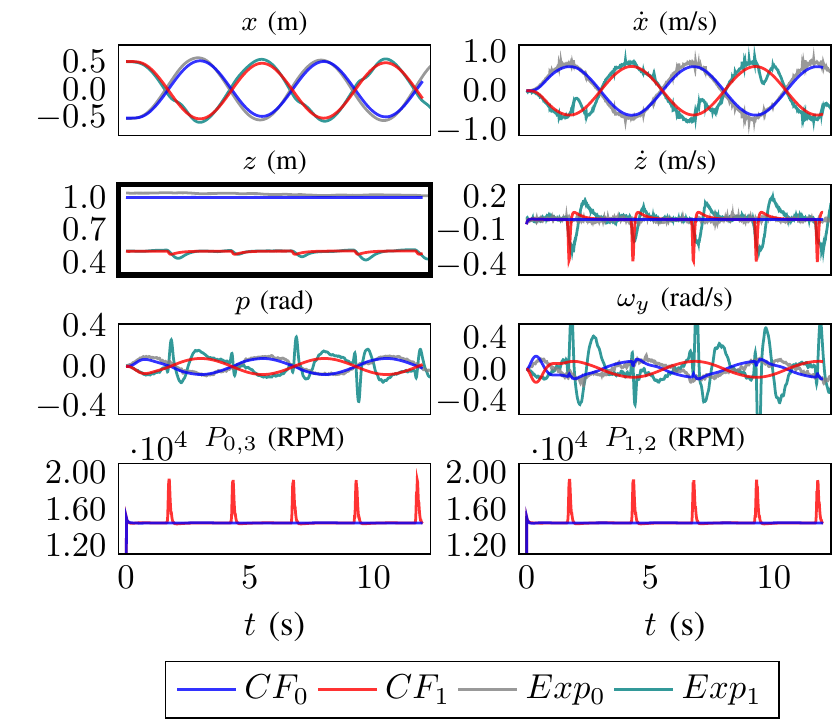}
    \caption{
    Positions in $x$ and $z$, linear velocities $\dot{x}$, $\dot{z}$, pitch $p$, angular velocity $\omega_y$, and motors' speeds $P_0 = P_3$, $P_1 = P_2$ of two Crazyflies $CF_0$ (blue), $CF_1$ (red) subject to the downwash model in~\eqref{eq:down}, compared to the flight logs of a real-world experiment with two drones (grey and teal lines).
    }
    \label{fig:down}
\end{figure}

The last control example subjects two Crazyflies---moving in opposite directions along sinusoidal trajectories in $x$ and different altitudes of 0.5 and 1 meter---to the downwash model in~\eqref{eq:down}.
\begin{lstlisting}[firstnumber=1,language=Python, 
    numbers=none,
label = {alg:down},]
$ python3 downwash.py --gui False
\end{lstlisting}
Figure~\ref{fig:down} compares the simulation results with the experimental data collected
to identify the parameters used in~\eqref{eq:down}.
Figure~\ref{fig:down} shows a very close match in the $x$ and $z$ positions between our simulation and the real-world. 

As we would expect, however, our simplified single contribution modeling does not fully capture the impact of downwash on the bottom drone---e.g., in the pitch $p$ and its ramifications on velocity $\dot{x}$.

\subsection{Reinforcement Learning}

The last two examples let one or more quadcopters learn policies to reach a target altitude and hover in place.
These are based on \emph{normalized} kinematics observations~\eqref{eq:obs}
and a \emph{normalized}, one-dimensional RPMs action space~\eqref{eq:rpms}.
\begin{lstlisting}[firstnumber=1,language=Python, 
    numbers=none,
label = {alg:rl},]
$ cd gym-pybullet-drones/experiments/learning/
\end{lstlisting}

\subsubsection{Single Agent Take-off and Hover}

\begin{figure}[t]
    \centering
    \includegraphics[]{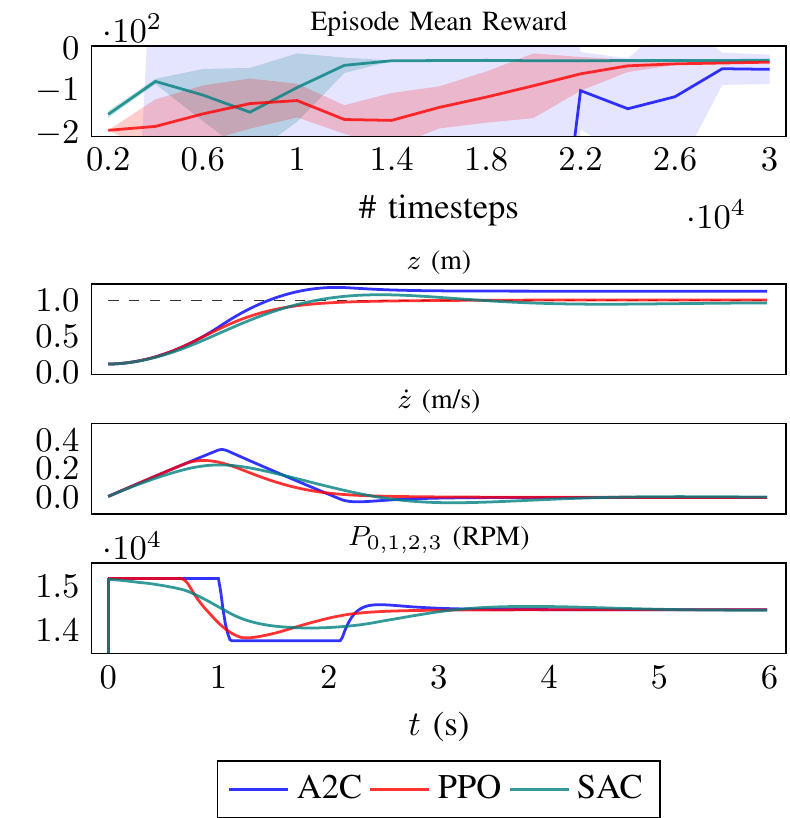}
    \caption{
    Algorithm's learning curve (top) and best policy's position in $z$, linear velocity $\dot{z}$, and motors' speeds $P_0 = P_1 = P_2 = P_3$ (bottom) for three single agent RL implementations from Stable Baselines3---A2C, PPO, and SAC---using the reward function in~\eqref{eq:single}.
    }
    \label{fig:single}
\end{figure}

For a single agent, the goal is to reach a predetermined altitude and stabilize. The reward function is simply the negation of the squared Euclidean distance from the set point:
\begin{equation}
    r = - \lVert [0,0,1] - \mathbf{x} \rVert_2^2
    .
\label{eq:single}
\end{equation}
We use the default implementations of three popular RL algorithms (PPO, A2C, and SAC) provided in Stable Baselines3~\cite{stable-baselines3}.
We do not tune any of the hyperparameters. We choose MLP models with ReLU activation and 4 hidden layers with 512, 512, 256, and 128 units, respectively.
For PPO, the training workflow can be executed as follows:
\begin{lstlisting}[firstnumber=1,language=Python, 
    numbers=none,
label = {alg:single1},]
$ python3 singleagent.py --algo ppo
\end{lstlisting}
To replay the best trained model, execute the following script:
\begin{lstlisting}[firstnumber=1,language=Python, 
    numbers=none,
label = {alg:single2},]
$ python3 test_singleagent.py --exp ./results/save-<env>-<algo>-<obs>-<act>-<time_date>
\end{lstlisting}
Figure~\ref{fig:single} compares \emph{(i)} the three algorithms' learning and \emph{(ii)} the trained policies performance.
While SAC performs best, all algorithms succeed albeit with very different learning curves. These were not unexpected, as we deliberately omitted any parameter tuning to avoid cherry-picking.

\subsubsection{Multi-agent Leader-follower}

\begin{figure}[t]
    \centering
    \includegraphics[]{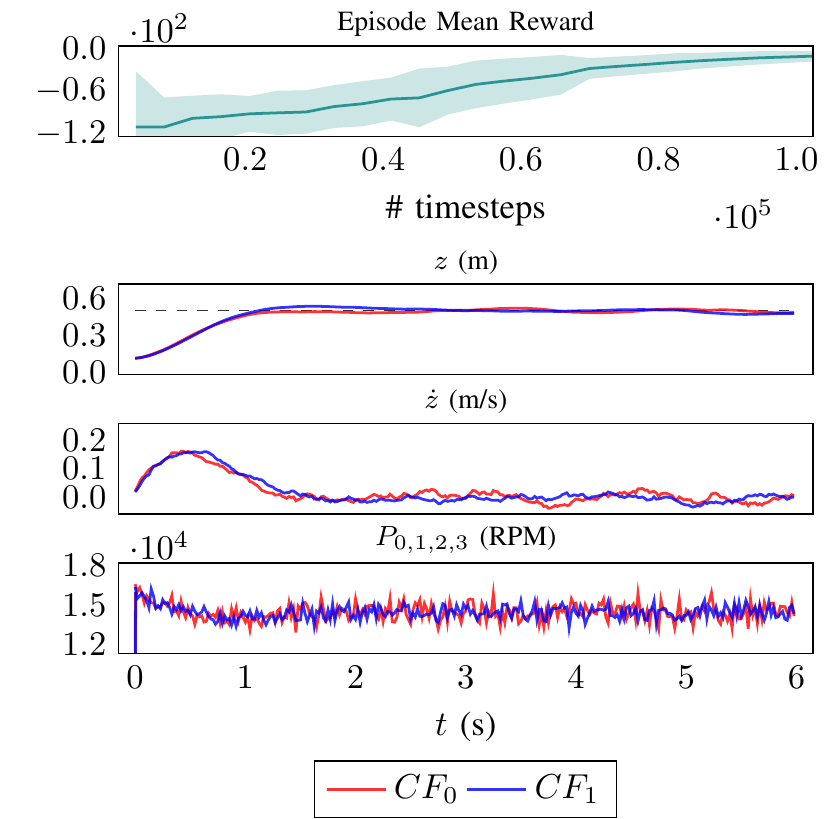}
    \caption{
    Learning curve (top) and best policies' positions in $z$, linear velocities $\dot{z}$, and motors' speeds $P_0 = P_1 = P_2 = P_3$ (bottom) for a 2-agent MARL implementation using RLlib and the reward functions in~\eqref{eq:multi}.
    }
    \label{fig:multi}
\end{figure}

The last example is a MARL problem in which a \emph{leader} agent is trained as in~\eqref{eq:single} and a \emph{follower} is rewarded by tracking its altitude: 
\begin{equation}
    r_0 = - \lVert [0,0,0.5] - \mathbf{x}_0 \rVert_2^2
    , \quad \quad
r_1 = - 0.5(z_1 - z_0)^2
.
\label{eq:multi}
\end{equation}
The workflow is built on top of RLlib~\cite{liang2018} using a central critic with 25 inputs and two action models with 12 inputs, all having two hidden layers of size 256 and tanh activations.
\begin{lstlisting}[firstnumber=1,language=Python, 
    numbers=none,
label = {alg:multi1},]
$ python3 multiagent.py --act one_d_rpm
\end{lstlisting}
To replay the best trained model, execute the following script:
\begin{lstlisting}[firstnumber=1,language=Python, 
    numbers=none,
label = {alg:multi2},]
$ python3 test_multiagent.py --exp ./results/save-<env>-2-cc-<obs>-<act>-<time_date>
\end{lstlisting}
Figure~\ref{fig:multi} shows a stable training leading to successfully trained policies. The leader presents minor oscillations that are, expectedly, reflected by the follower. The RPMs commanded by these policies, however, appear to be erratic.

 \section{Extensions}
\label{sec:future}

We developed \texttt{\small gym-pybullet-drones} to provide a compact and comprehensive set of features 
to kick-start RL in quadcopter control.
Yet, we also structured its code base for extensibility.
Some of the enhancements in the works include:
\emph{(i)} the support for heterogeneous quadcopter teams---this can be achieved by importing multiple URDF files, where the inertial properties are stored;
\emph{(ii)} the development of more sophisticated aerodynamic effects---e.g., a downwash model 
made of multiple components instead of a single
contribution applied to the center of mass;
\emph{(iii)} the inclusion of symbolic 
dynamics---e.g., using CasADi to expose an analytical model that could be
leveraged by model predictive control approaches;
\emph{(iv)} new workflows to support additional MARL frameworks beyond RLlib~\cite{liang2018}---e.g., PyMARL; and finally,
\emph{(v)} Google Colaboratory support and Jupyter Notebook examples---to facilitate adoption by those with limited access to computing resources.

\section{Conclusions}
\label{sec:conclusions}

In this paper, we presented an open-source, OpenAI \emph{Gym}-like~\cite{brockman2016} multi-quadcopter simulator written in Python on top of the Bullet Physics engine~\cite{coumans2019}.
When compared to similar existing tools, 
the distinguishing and innovative features of our
proposal include \emph{(i)} a more modular and sophisticated physics implementation, \emph{(ii)}
vision-based \emph{Gym}'s observations, and
\emph{(iii)} a multi-agent reinforcement learning interface.
We showed how \texttt{\small gym-pybullet-drones}
can be used for low- and high-level control
through trajectory tracking and target velocity
input examples. 
We also demonstrated the use of our work 
in separate workflows for single and multi-agent RL,
based on state-of-the-art learning libraries~\cite{stable-baselines3,liang2018}.
We believe our work will contribute to 
bridging the gap between reinforcement learning and control research, helping the community to develop realistic MARL applications for aerial robotics.

\section*{Acknowledgments}

We acknowledge the support of Mitacs's Elevate Fellowship program and General Dynamics Land Systems-Canada (GDLS-C)'s Innovation Cell. 
We also thank the Vector Institute for providing access to its computing resources.

\balance

\balance
\end{document}